\documentclass[authoryear,11pt,square,number,times,super,comma,3p]{article}

\usepackage[authormarkup=none,markup=underlined]{changes}

\usepackage{amstext,amsmath,amssymb,amsthm}
\usepackage{algorithm}
\usepackage{endnotes}
\usepackage{algpseudocode}
\usepackage{xcolor}
\usepackage{hyperref}
\usepackage{rotating}
\usepackage{caption}
\usepackage{natbib}
\usepackage[a4paper, total={170mm,257mm},
 left=20mm,
 top=20mm]{geometry}

\newcommand{\NULL}{\textsc{null}}
 \newcommand{\Tau}{\mathcal{T}}
 \newcommand{\pib}{S^{ib}}
 
 \newcommand{\piib}{\pi^{ib}}
 
 \newcommand{\pibs}{\pi^{bs}}
 \newcommand{\pbs}{S^{bs}}

 \DeclareMathOperator*{\argmax}{\arg\!\max}
 \DeclareMathOperator*{\argmin}{arg\,min}

\begin{document}

\title{An Efficient Merge Search Matheuristic for Maximising the Net Present Value of Project Schedules}

\author{Dhananjay R. Thiruvady$^a$,  Su Nguyen$^b$, Christian Blum$^c$, Andreas T. Ernst$^d$ \\ \\
$^a${\it School of Information Technology, Deakin University, Geelong, Australia}
 \\
$^b${\it Centre for Data Analytics and Cognition, La Trobe University, Australia}
 \\
$^c${\it Artificial Intelligence Research Institute (IIIA-CSIC), Bellaterra, Spain}
 \\
$^d${\it School of Mathematical Sciences, Monash University, Australia}}

\maketitle

\begin{abstract}
Resource constrained project scheduling is an important combinatorial optimisation problem with many practical applications. With complex requirements such as precedence constraints, limited resources, and finance-based objectives, finding optimal solutions for large problem instances is very challenging even with well-customised meta-heuristics and matheuristics. To address this challenge, we propose a new math-heuristic algorithm based on Merge Search and parallel computing to solve the resource constrained project scheduling with the aim of maximising the net present value. This paper presents a novel matheuristic framework designed for resource constrained project scheduling, Merge search, which is a variable partitioning and merging mechanism to formulate restricted mixed integer programs with the aim of improving an existing pool of solutions. The solution pool is obtained via a customised parallel ant colony optimisation algorithm, which is also capable of generating high quality solutions on its own. The experimental results show that the proposed method outperforms the current state-of-the-art algorithms on known benchmark problem instances. Further analyses also demonstrate that the proposed algorithm is substantially more efficient compared to its counterparts in respect to its convergence properties when considering multiple cores.
\end{abstract}

\section{Introduction}\label{sec:intro}
Project scheduling has been a problem of interest for many years. There are a number of variants, but the main aspect of the problem is to complete several tasks with an objective associated with the total completion time or, in other words, the cumulative value of tasks. The project scheduling problem has a number of variants~\citep{bruker99}. The common aspects of all of these variants are that the projects consist of tasks, shared renewable resources (with limits) and a deadline. There might be precedence constraints between the tasks. Moreover, tasks use some proportion of the available resources. To solve these problems, several different methods have been proposed. These include heuristics, meta-heuristics (simulated annealing, genetic algorithms, tabu search, etc.), and branch \& bound. A number of these approaches (heuristics, meta-heuristics and exact techniques) and their associated details were discussed by~\citet{demeulemeester02} and~\citet{Neuman03}. Furthermore, \citet{Neuman03} discuss heuristic and exact approaches for project scheduling with time windows.

An important variant of project scheduling is the resource constrained project scheduling (RCPS) with the \emph{net present value} (NPV) objective (RCPS-NPV)~\citep{kimms00}. In this problem, all tasks are associated with a cash flow, either positive or negative, and the goal is to maximise the NPV, i.e. the sum of the discounted cash flows of the tasks. This problem has received a lot of attention recently as it reflects the financial health or feasibility of a project. RCPS-NPV is a very complex problem for which several different approaches have been proposed. \citet{Chen10} investigate the problem with 98 tasks using ant colony optimisation (ACO), outperforming other meta-heuristic methods including simulated annealing, genetic algorithms and tabu search. \citet{Show06} also investigate the use of ACO and show that their approach is effective for problem instances with up to 50 tasks. \citet{Vanhoucke2010} propose a scatter search heuristic, which outperforms a branch \& bound method previously developed by the same authors for the same problem \citep{Vanhouke2001}. \citet{Gu2013} suggest the use of constraint programming within a Lagrangian relaxation framework in order to find high-quality feasible solutions to the problem. In a similar fashion, \citet{Thiruvady2014ps} developed a hybrid of Lagrangian relaxation and ACO, which was shown to be very effective, particularly when implemented in a parallel setting \citep{Brent14}. 

Construct, solve, merge and adapt (CMSA) is a method that was recently introduced and applied to a number of combinatorial optimisation problems \citep{Blum2016,BlumBlesa16,Blum16-2,Lewis2019,Polyakovskiy2020, ThiruvadyBlumErnst2020,Thiruvady2022}. \citet{Blum2016} introduced this method and applied it to two problems, namely the minimum common string partition problem and the minimum covering arborescence problem. \citet{BlumBlesa16} then showed that CMSA is effective on the repetition-free longest common subsequence problem, and \citet{Blum16-2} showed the same for the unbalanced common string partition problem. The CMSA method works as follows. There are four stages to the algorithm: (a) generate a number of `good' solutions and add the components found in these solutions to a pool of solution components, (b) identify a restricted mixed integer program (MIP) based on the pool of solution components, (c) solve the MIP to generate a `merged' solution, and (d) update the pool of solution components by incorporating information from the MIP. The above stages are iteratively applied until some termination criteria are reached. \citet{Thiruvady2019} proposed a hybrid of CMSA and parallel ACO for RCPS-NPV, which produces the currently best results in the literature. \citet{Polyakovskiy2020} apply CMSA to a just-in-time batch scheduling problem, with promising results.  The study by \citet{ThiruvadyBlumErnst2020} show that CMSA is effective in tackling a resource constrained job scheduling problem.  Most recently, \citet{Thiruvady2022}    combine CMSA with Tabu search for the maximum happy vertices problem, and they find that the hybrid produces excellent results.  

Merge search is a very recent technique proposed for a range of scheduling problems \citep{Kenny:2018}. The algorithm is similar to that of CMSA, where presumably good solutions are used to generate a restricted MIP, which in turn, is solved to obtain a solution that is used as the input to generate a new pool of solutions. The key differences are in how the restricted MIP is generated and how its solution space may be efficiently improved to incorporate more diversity (via a random splitting mechanism). The restricted MIP in Merge search is generated by an aggregation of variables (details in Section~\ref{sec:methods}), depending on a pool of solutions. Random splitting is now applied to the aggregated MIP, where the aggregated variable sets are split to allow a larger neighbourhood to be searched. Relatively few studies have been conducted with Merge search, but the existing ones shown excellent results. \citet{Kenny:2018} showed that Merge search is very effective for the constrained pit problem.

Merge search has two key advantages over conventional methods such as genetic algorithms \citep{Mitchell:1998}. Firstly, it is able to manage the trade-off of solving a straightforward MIP (via variable aggregation) but yet consider a substantially diverse search space obtained by merging solution components from a very large population of solutions. Secondly, the method is inherently parallelisable, lending itself to multi-core shared memory architectures~\citep{Dagum98} or the message passing interface (MPI) \citep{Gropp:1994}. Indeed, parallel ACO (PACO)~\citep{Thiruvady2016} and a parallel hybrid of ACO and constraint programming~\citep{Cohen2017} has shown to be very effective on a related resource constrained job scheduling problem. By exploiting the relative advantages of Merge search and PACO, this study tackles the RCPS-NPV, thereby further validating this technique for scheduling.

In this study we propose an efficient Merge search algorithm with PACO (MS-PACO) for tackling the RCPS-NPV problem. An efficient practical implementation of MS-PACO is achieved through a tight coupling between the MIP component and PACO, where PACO provides good areas of the search space for the MIP to explore, and in turn, the MIP provides improvements which are used in the learning component of PACO. In particular, this framework is efficient because (i) a solution pool with large diversity can be obtained quickly, (ii) the search space can be enlarged efficiently via random splitting and (iii) restricted MIPs can be solved in short time-frames. We demonstrate the efficacy of MS-PACO on a large number of RCPS-NPV instances, where it outperforms previous algorithms, especially the most recent one involving CMSA and PACO \citep{Thiruvady2019}. Overall, the contributions of this study can be summarised as follows:
\begin{itemize}
    \item Regarding PACO: we provide an efficient implementation of PACO, which is very effective at generating a set of diverse and quality solutions.
    \item Regarding the combination of Merge search with PACO: we introduce a new variable partitioning and merging mechanism to formulate restricted MIPs to further improve solutions for the RCPS-NPV.
    \item We present an efficient way to utilise the high-quality solutions found by PACO to improve them via a restricted MIP.
    \item Regarding results: our algorithm obtains improvements of best-known-solutions for numerous RCPS-NPV benchmark instances.
\end{itemize}

The paper is organized as follows. Section~\ref{sec:rcp} formally discusses the RCPS-NPV problem and provides an efficient  MIP model. Section~\ref{sec:methods} discusses Merge search and PACO, including the intuition behind the methods and their integration. Section~\ref{sec:results} details the experiments conducted and the results obtained from these experiments. This includes an analysis of the algorithms on known benchmark problem instances and an analysis of the algorithm's convergence characteristics. Section~\ref{sec:conclusion} concludes the paper and provides potential future direction.

\section{Formulating the RCPS-NPV Problem}\label{sec:rcp}

The formal definition of the RCPS-NPV problem is as follows. Given is a set ${\cal J}$ containing $n$ tasks. Each task $i \in {\cal J}$ has a duration $d_i$, a resource consumption $r_{ik}$ for each of $k$ resources, and a cash flow $cf_{it}$, which depends on the  time period $t \geq 1$. The cash flows may be positive or negative. Moreover, the cash flow of a task may vary over its duration. Its total net value $c_i$, however, can be calculated if it would be completed at time point $0$. Using this value we can determine a discounted value for future time points by applying a discount factor $\alpha>0$ to the start time $s_i$ of the task. We use $c_i\, e^{-\alpha(s_i+d_i)}$ \citep{kimms00}, where $d_i$ is task $i$'s duration. Note that the discount formula $e^{-\alpha\,t}$ is equivalent to the commonly used function $1/(1+\bar\alpha)^t$ for some $\bar\alpha$. We are also given a set of precedence constraints ${\cal P}$, where $i\rightarrow j$ or $(i,j)\in {\cal P}\subseteq \mathcal{J}\times\mathcal{J}$ represents that task $i$ needs to complete before $j$ starts. 

For the $k$ resources we are given limits $R_1, \ldots, R_k$, and the cumulative use of the resources by the tasks at all time points must satisfy these limits. Finally, we are given $\delta$, which is the deadline for all tasks and, therefore, the time horizon to complete the project. This deadline ensures that tasks with a negative cash flow and no successors will still be completed. The objective is to maximise the NPV and the problem can then formally be stated as follows:
 \begin{alignat}{3}
 &&\max\quad& \sum_{i \in \cal J} c_i\,e^{-\alpha(s_i+d_i)} & \label{obj}\\
\text{S.T.}\quad&&  s_i + d_i &\,\leq\, s_j &&\forall\ (i,j) \in P \label{c1} \\ 
&& \sum_{i \in S(t)} r_{im} &\,\leq\, R_m &&\forall\ m = 1, \ldots k \label{c2} \\ 
&& 0\,\leq\, s_i &\,\leq\, \delta - d_i && \forall\ i \in {\cal J}\label{c3}
\end{alignat}

$S(t)$ is the set of tasks executing at time $t$. Function~\eqref{obj} is the NPV objective, which is very challenging to solve as it is non-linear and neither convex nor concave. Constraints~\eqref{c1} ensure that the precedence constraints are satisfied. Constraints~\eqref{c2} enforce the resource limits and Constraints~\eqref{c3} ensure that the deadline is not violated.

Similar to the studies by \citet{kimms00} and \citet{Thiruvady2014ps}, we will consider deadlines that are relatively loose, since minimizing the makespan is not an objective. This results in sufficient slack which allows negative-valued tasks to be scheduled later, thereby increasing the NPV slightly.

\subsection{An Integer Programming Model}\label{sec:IPmodel}

An integer programming (IP) model for the RCPS-NPV problem is given by \citet{kimms00}. However, the study by \citet{Thiruvady2014ps} showed that this model can be made computationally more efficient. Hence, we use the latter model as the basis for this study. 

The improved model can be defined as follows: Let $x_{it}$ be a binary variable for each task $i \in \cal J$ and time point $t \in \{1,\ldots,\delta\}$. $x_{it}$ takes value 1 if task $i$ has already completed by time point $t$. The MIP model can then be stated as follows: 
\begin{equation}
 \max \quad \sum_{i \in \cal J} \sum_{t=2}^{\delta} c_i\,e^{-\alpha t}\,(x_{it} - x_{it-1})  \label{m_obj}
\end{equation}
Subject to
 \begin{alignat}2
 &x_{it} \,\geq\, x_{it-1} & & \forall\ i \in {\cal J},  t \in \{2,\ldots,\delta\} \label{m_c1} \\ 
 &  x_{i\delta} \,= 1  & &\forall\ i \in \cal J \label{m_c2}\\
 & x_{i,t} \,=0 &&\forall\ i\in \mathcal{J},\ t\in\{1,\ldots,d_j-1\}\\
&  x_{jt} \, \leq x_{i,t - d_j} &  & \forall\ (i,j)  \in {\cal P},  t \in \{d_j+1,\ldots,\delta\} \label{m_c3}\\ 
& \sum_{i\in\mathcal{J}:\atop d_i>t}  r_{ik}\, (x_{it} - x_{i,t-d_i}) \,  &&\leq\, R_{mt}  \nonumber
\quad\forall\ m =  \{1,\ldots,k\},\\[-5mm] 
&&&\hphantom{\leq\, R_{mt}\ \quad\forall\ } t \in \{1,\ldots,\delta\}  \label{m_c4} \\ 
& x_{it} \,\in\, \{0,1\}  &  & \forall\ i \in {\cal J},  t \in \{1,\ldots,\delta\} \label{m_c5} 
\end{alignat}

Equation~\eqref{m_obj} is the objective, aiming to maximise the NPV. Constraints~\eqref{m_c1} ensure that once a task is completed it stays completed. Constraints~\eqref{m_c2} ensure that all tasks complete. Constraints~\eqref{m_c3} ensure that all precedence constraints are satisfied. Constraints~\eqref{m_c4} require that, at each time point, none of the resource limits are exceeded. Finally, the variables $x_{it}$ are binary. 

This formulation often results in a high density of variables with one-zero values. As we will shortly see, this proves extremely beneficial when applying MS-PACO to this problem. We also note that for MS-PACO we never solve the complete model. Instead, by significantly restricting the set of feasible time points, we solve substantially reduced subproblems. In preliminary experimentation we found that such MIPs can be easily solved with a general-purpose MIP solver.

\section{The Proposed Algorithm} \label{sec:methods}

In this section, we focus on MS-PACO and its particular implementation for the RCPS-NPV. Using an initial pool of (good) solutions, this algorithm generates an aggregated MIP, that is efficiently solved by a general-purpose MIP solver. The initial solutions themselves are found by running multiple colonies of ACO in a parallel setting, as discussed in~\citep{Thiruvady2012}. Solutions in future iterations are also found by this method.

We first discuss the parallel ACO (PACO) approach. Compared to previous multi-threaded implementations of ACO for this and similar resource constrained scheduling problems, we implement a multi-colony ACO. In this approach the colonies share---at regular intervals---information on their best solution. This approach, in its own right, proves effective on small problem instances (see Section~\ref{sec:results}).

This is followed by a discussion of Merge Search and its integration with PACO. As previously discussed, Merge Search requires a population of solutions which are of high quality and which are characterised by sufficient diversity. A key requirement for the generation of this population is that this is done efficiently, and hence, PACO provides an ideal way to achieve this. This population is used to build an aggregated model or restricted MIP, which can be solved very efficiently compared to the original MIP. The outputs of the MIP are passed back into PACO, which is used to build a new population. Our results show that this approach is indeed very effective for the RCPS-NPV problem, achieving best results for a number of problem instances (see Section~\ref{sec:results}).    

\subsection{Ant Colony Optimisation and its Parallel Implementation}\label{sec:paco}

ACO is a method for solving combinatorial optimisation problems which was first proposed by~\citet{dorigo92}. These algorithms are inspired by the foraging behaviour of real ants, where short paths between nests and food sources are found through positive reinforcement. This technique has shown to be effective on a number of problems \citep{dorigo04}. Project scheduling variants have also been solved with ACO \citep{Merkle00,Chen10,Show06}. The RCPS problem with makespan as the objective was considered, for example, by \citet{Merkle00}. Moreover, an ACO approach for project scheduling where tasks can execute in multiple modes is described in \citep{Chen10}. In the study by \citet{Show06}, a problem similar to the one studied by \citet{Vanhoucke2010} was tackled, albeit with fewer tasks.

\citet{Thiruvady2014ps} propose an ant colony system (ACS) approach for the RCJS-NPV problem. For the same problem, \citet{Brent14} extend ACS to a parallel implementation using a multi-core shared memory architeture. The recent study by \citet{Thiruvady2019} used parallel ACS with multiple colonies to generate a solution pool within CMSA. In this study, we use this approach within Merge Search. Furthermore, in this study, we extend this approach to consider parallel ACS as an independent algorithm for which we allow information sharing between the colonies. We provide a comprehensive description of the approach here, but further details can in found in \citep{Thiruvady2019}. \\ 

Algorithm~\ref{alg:acs_cs} shows the outline of the ACS procedure. The inputs to the algorithm are the pheromone trails $\Tau$ and a solution. A solution ($\pibs$), typically the best solution found so far by the overall approach, can be provided as input to bias the search. If no solution is provided as input, the algorithm makes use of a random starting point. The main part of the algorithm is executed between Lines~3 and~13. The algorithm runs until either a time limit or an iteration limit is reached. \\

\begin{algorithm}[tb]
\caption{ACS for the RCPS-NPV problem}
\label{alg:acs_cs}
\begin{algorithmic}[1]
  \State {\bf input}: An RCPS-NPV instance, $\Tau$, $\pibs$ (optional)
  \State Initialise $\pibs$ (if given as input, otherwise not)
\While{termination conditions not satisfied}
  \For {$j = 1$ to $n_{\mathrm{ants}}$}:
  	\State $\pi^j :=$ {\sf ConstructPermutation($\Tau$)}
    \State {\sf ScheduleTasks($\pi^j$)}
   \EndFor
  \State $\piib := \argmin_{j=1,\ldots,n_{\mathrm{ants}}} f(\pi^j)$
  \State $\pibs :=$ {\sf Update($\piib$)}
  \State {\sf PheromoneUpdate}($\Tau$, $\pibs$)
  \State $cf :=$ {\sf ComputeConvergence}($\Tau$)
  \State {\bf if} {\it cf} = true {\bf then} {\it initialise} $\Tau$ {\bf end if}
\EndWhile
\State {\bf output:} $\pibs$ (converted into a CMSA solution)
\end{algorithmic}
\end{algorithm}

\subsubsection{Solution representation and pheromone model}
In the ACS model, solutions are represented in terms of permutations $\pi$ of the tasks. This is in contrast to the MIP model, which focuses on the completion times of the tasks. The main reason for choosing permutations for representing solutions in ACS is that this results in a much smaller pheromone model than with the other option of directly learning the starting times of the tasks. Once we have a permutation of all tasks, a scheduling heuristic can be used to generate a schedule---that is, the tasks' starting times---in a deterministic way. As previous studies have shown, forward-backward scheduling heuristics have been very effective for this purpose~\citep{Gu2013}. We describe some of these approaches in Section~\ref{sec:listSched} below. At this stage it is sufficient to know that the restricted MIP model of MS-PACS can be generated from an ACS solution in an efficient manner. In this context, note that $f(\pi)$ denotes the objective function value of the solution generated by the scheduling heuristic on the basis of $\pi$. 

We use a pheromone model that is similar to the one used in~\citep{basten00}. Here, the pheromone trails ($\Tau$) consist for each position $i$ in $\pi$ and for each task $j \in {\cal J}$ of a parameter $\tau_{ij}$. Hereby, $\tau_{ij}$ is the desirability of picking task $j$ for position $i$ of a permutation. \\

\subsubsection{Construction permutations}
At each iteration, $n_{\mathrm{ants}}$ solutions are built from scratch by function {\sf ConstructPermutation($\Tau$)}. Each of these permutations $\pi$ is incrementally built by selecting one task per step using the pheromone trails $\Tau$. By $\hat{\cal J}$ we denote the set of tasks that can be potentially chosen for position $i$. That is, $\hat{\cal J}$ is the set of tasks whose predecessors have already been scheduled and that have not been assigned to $\pi$ yet. The selection of a task happens in one of two ways. First, $q \in (0,1]$, a random number, is generated and if $q < q_0$, the task is selected deterministically: 
\begin{equation} \label{eq:sampling_1}
  k = \argmax_{j \in \hat{\cal J}} \tau_{ij}
\end{equation}
However, if $q \geq q_0$, task $k$ is selected probabilistically by roulette wheel selection according to the following probabilities:
\begin{equation} \label{eq:sampling}
  P(\pi_i=l) = \frac{\tau_{il}}
  		{\mbox{$\sum_{j\in \hat{\cal J} }$}\,\tau_{ij}} \quad, l \in \hat{\cal J}
\end{equation}
Subsequently, task $k$ is placed at position $i$ of $\pi$. \\

\subsubsection{Schedule construction}\label{sec:listSched}
We have pointed out earlier that---on the basis of a permutation $\pi$---a feasible schedule respecting all the precedence and resource constraints can be obtained in a determinimistic manner. We discuss this method here at a high-level. For further details we refer the reader to \citep{Thiruvady2014ps}.

We start with $\pi$, a complete permutation of the tasks. In the order defined by $\pi$, sets of tasks are chosen where the task and its chain of successors are included. That is, the set consists of a task, its immediate successor, the successors of these successors, and so on, until no successor remains. For the chosen set of tasks, we can determine if it is net positive-valued or net negative-valued. That is, all the tasks in the set together have a total NPV which is either negative or positive. The sets which are positive-valued are scheduled starting as early as possible (in a greedy fashion), ensuring that the resource constraints are satisfied. Similarly, the set of negative-valued tasks are scheduled in order but starting at the end of the time horizon and working backwards. This way of scheduling the tasks allows positive-valued tasks to be scheduled early, thereby increasing the NPV, and negative-valued tasks towards the end of the horizon, thereby also increasing the NPV.

When dealing with a given set of positive-valued tasks, the procedure ensures that precedence constraints are satisfied. Tasks are examined in the order they appear in the original permutation. A task, in order, is selected to be scheduled and examined if it is precedence feasible (i.e. all its preceding tasks have been completed). If not, the task is task is ignored and the process continues in order to find a precedence feasible task. Once a task is scheduled, the set is again examined in order to see if any other tasks are now precedence feasible as a result of the scheduled task. Once identified, the new task is immediately scheduled. \\

\subsubsection{Pheromone updating mechanism}
Our ACS algorithm makes use of two different pheromone update mechanisms. The first one is applied during solution construction. Whenever a task $k$ is selected for position $i$, the following update is applied:
\begin{equation}
\tau_{ik} := \tau_{ik} \times \rho + \tau_{min}
\end{equation}
To ensure a positive lower bound for all pheromone values---such that any task has always a non-zero probability of being selected for any position---we set $\tau_{min} := 0.001$.

At each iteration, after the construction of $n_{\mathrm{ants}}$ solutions, the best of these is selected to be the iteration-best solution $\piib$ (Line~8). Following this, $\pibs$ or the best-so-far solution is updated in {\sf Update($\piib$)}, where a new best-so-far solution is chosen if $f(\piib) > f(\pibs) \Rightarrow \pibs := \piib$. Afterwards, the second type of pheromone update is applied in function {\sf PheromoneUpdate}($\pibs$). In particular, the pheromone values corresponding to the solution components in $\pibs$ are updated as follows:
\begin{equation}
\tau_{i\pibs(i)} = \tau_{i\pibs(i)} \cdot \rho + \delta \quad,
\end{equation}
where a small constant reward $\delta = 0.01$ is used. The evaporation rate, $\rho = 0.1$, is set to be a relatively high value, which was also used in the previous study~\citep{Thiruvady2014ps}. The motivation for using a high value is since the ASC procedure will be given a short time-frame, respectively, a low number of iterations to run within MS-PACO. Only very little time can be spent on this part of the algorithm, and hence, a high learning rate allows converging to local optima more quickly. \\

\subsubsection{The parallel implementation}

\begin{figure}[t!]
  \includegraphics[width=17cm]{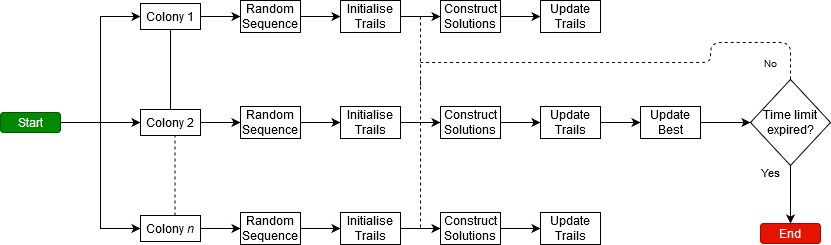}
  \centering
  \caption{The flowchart for PACS. Multiple colonies execute in parallel and they synchronise when their best solutions are compared to determine which ones are best.}
  \label{fig:paco}
\end{figure}

As a final note, \citet{Brent14} showed that ACS can be parallelised in different ways. The most obvious choice is to build the solutions concurrently, accounting for the local pheromone update. In this study we chose not to use this specific parallel implementation. Instead we use the one studied in \citep{Thiruvady2019}. The motivation for this revised scheme is to achieve increased diversity. Essentially, the parallelism consists of multiple colonies rather than multiple solutions constructed in parallel. The reasons for using this approach are twofold. Firstly, we observe that there is a need for multiple, potentially unique, solutions. If we use a single colony and select a number of the best solutions, we find they they contain many overlapping solution components. This is because these solutions are from the same region of the search space. Hence, the diversity needed to generate a good restricted MIPs seems insufficient. A second, and more minor reason, is that when constructing solutions in parallel, the utilisation of cores may not be very high since the algorithm also spends some time in the serial parts of the algorithm, such as the pheromone update. Therefore, our MS-PACS algorithm runs $n_s$ colonies in parallel, ensuring sufficient diversity and 100$\%$ utilisation of the cores. \footnote{Note that in order to achieve diversity, we have to have many colonies seeded randomly.}

The flow of this approach can be seen in Figure~\ref{fig:paco}. The main difference between the different colonies is the starting point (random sequences). The colonies are all independent until the point of updating the common best-so-far solution.

\subsection{Merge Search with parallel ACS} \label{sec:cmsa}

Algorithm~\ref{alg:ms} presents the MS-PACS algorithm for the RCPS-NPV problem. The inputs to the algorithm are (1) an RCPS-NPV problem instance, (2) a number of solutions to be generated at each iteration - $n_s$, (3) a time limit - $t_{\mathrm{total}}$, (4) a time limit per MIP to be solved at each iteration - $t_{\mathrm{iter}}$, and (5) a number of sets to split on - $K$. A solution $S$ is a feasible solution to the RCPS-NPV problem and its objective value will be denoted by $f(S)$.

\begin{algorithm}[ht!]
\caption{MS-PACS for the RCPS-NPV problem}
\label{alg:ms}
\begin{algorithmic}[1]
\State {\bf \sc input:} RCPS-NPV instance,  $n_s$, $t_{total}$, $t_{iter}$, $K$
\State Initialisation: $\pbs := \NULL$
\While {time limit $t_{\mathrm{total}}$ not expired}
  \State {\bf if} $\pbs \not= \NULL$ {\bf then} ${\cal S} := \{\pbs\}$ {\bf else} ${\cal S} := \emptyset$ {\bf end if}
  \For{$i=1,2,\ldots,n_s$} \hfill \# this is done in parallel
    \State $S :=$ {\sf GenerateSolution($\pbs$)}
    \State ${\cal S} \leftarrow {\cal S} \cup S$
  \EndFor
  \State ${\cal P}$ $:=$ {\sf Partition(${\cal S}$)} 
\State ${\cal P}'$ $:=$ {\sf RandomSplit(${\cal P}$, $K$)} 
  \State $\pib :=$  {\sf Apply\_MIP\_Solver(${\cal P}'$,$\pbs$,$t_{iter}$)}
  \State {\bf if} $\pbs = \NULL$ {\bf or} $f(\pib) < f(\pbs)$ {\bf then} $\pbs := \pib$ {\bf end if}
\EndWhile
\State {\bf \sc output:}$\pbs$
\end{algorithmic}
\end{algorithm}

There are four main steps in Algorithm~\ref{alg:ms}, which are repeated until a terminating criterion is met. However, before these steps are performed, a data structure to represent the best-so-far solution - $\pbs$ - is initialised. The main part of the MS-PACS algorithm (Lines~3--13) executes until a time limit $t_{\mathrm{total}}$ is reached. \\

\subsubsection{Generating a solution pool}
In Line~6, a number of solutions ($n_s$) is generated with a bias of the best-so-far solution {\sf GenerateSolution($\pbs$)}. Any method can be used to generate these solutions. However, in this work we make use of PACS (as outlined in Section~\ref{sec:paco}). As already mentioned, the generated solutions are required to be from good regions of the search space and must contain sufficient diversity. However, in contrast to the standalone application of PACS in which all parallel colonies are seeded by random solutions, when used within MS-PACS one of these colonies is seeded with the best-so-far solution $\pbs$. We found in preliminary experimentation that this method finds good solutions and at the same time provides substantial diversity. Only the best solution from each colony is included in the solution pool.\\

\subsubsection{Variable partitioning}
\begin{figure*}
  \includegraphics[width=12cm]{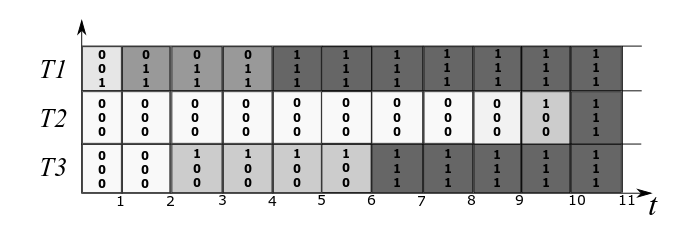}
  \centering
  \caption{An example problem instance for the RCPSP-NPV problem with three tasks (T1, T2, T3). The variable values of three solutions are shown (see the three rows of values for each task). For example, the first row for task T1 at time point 1 is 0 indicating that T1 is not yet completed. The first appearance of a 1 (with increasing time) in a row indicates the time at which the task completes. For example, in solution 1, task 1 completes at time point 5, while in solution 3 it completes at time point 1.}
  \label{fig:MSorig}
\end{figure*}

Once the pool of solutions {\cal S} has been generated, the variables of the MIP model are split into disjoint sets (Line~9). These sets depend on the start times of each task across all the solutions. An example is shown in Figure~\ref{fig:MSorig} with three tasks, 11 time points and three solutions. A set of variables consists of those variables that have the same values in all solutions of {\cal S}. The sets concerning the example are indicated in Figure~\ref{fig:MSorig} by different shades of gray concerning the background color. Note that there are 11 variables in the set determined by a value of 0 in all three solutions. \\

\subsubsection{Random splitting}
\begin{figure*}
  \includegraphics[width=12cm]{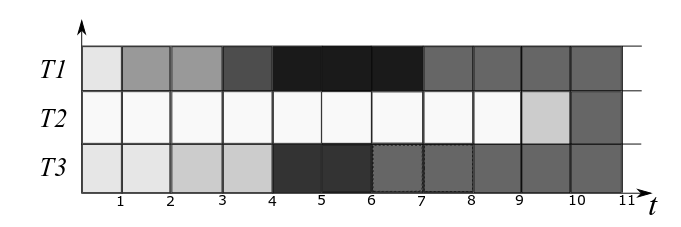}
  \centering
  \caption{Example for random splitting. Three of the variable sets from Figure~\ref{fig:MSorig} are further split, each one into two disjoint subsets. 
  }
  \label{fig:MS2}
\end{figure*}

In Line~10, the variable sets determined in the previous step are further split. This is to improve the diversity of potential solutions to be generated by the MIP model in Line~11. In this study we use random splitting, i.e., using a pre-defined parameter $K$, each set is split based on randomly chosen time points into $K$ non-empty subsets.\footnote{We note that the splitting may be applied taking into account additional information on the problem or gathered through the search process. However, we do not investigate this aspect in this study and leave it to future work.} If a variable set is too small, e.g. a single time point, no splitting is applied. An example of random splitting concerning the example in Figure~\ref{fig:MSorig} is shown in Figure~\ref{fig:MS2}. We see that, for instance, the set defined by value 1 in the first solution, value 0 in the second and third solutions, has four variables and is randomly split into two independent sets of size two. 
\\

\subsubsection{Merging mechanism}
The \emph{merge} phase of the algorithm is applied - {\sf Apply\_MIP\_Solver(${\cal P}'$,$\pbs$,$t_{iter}$)}. Here, a MIP model based on the disjoint sets of variables is defined and solved with time limit $t_{iter}$. The model itself is obtained from the full MIP model defined in Section~\ref{sec:IPmodel}. However, the variables are now aggregated according to the sets generated (detailed above). The resulting restricted MIP has generally much fewer variables and can be solved relatively easily with low computational time limits. However, if a lot of random splitting is used, the search space can grow potentially leading to very large computational times. Note that, in the extreme case---i.e., with complete random splitting---the original MIP model is obtained. The restricted MIP model is also seeded with the best-so-far solution to possibly enhance the performance of the MIP solver. ($t_{\mathit{iter}}$) is an imposed time limit in order to avoid that one iteration of the algorithm uses up all the available computation time. \\

In Line~12, if an improvement over the best-so-far solution is found, it is updated. Finally, at the end of the procedure, $\pbs$ is provided as output.

\subsection{The MS - PACS Hybrid}

\begin{figure}[t!]
  \includegraphics[width=17cm]{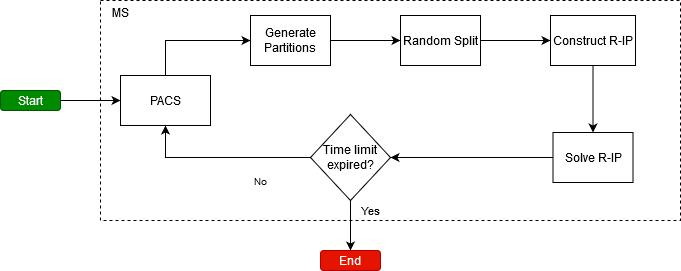}
  \centering
  \caption{The integration of Merge search and PACS. At each iteration, PACS is executed in parallel in order to generate $n_s$ solutions. These solutions (in addition to the best-so-far solution) are used to determine the disjoint sets/partitions of the variables. Afterwards, random splitting is applied. Then, the restricted MIP (R-MIP) is generated and then solved. This completes one iteration. The process continues until the algorithm terminates.}
  \label{fig:integration}
\end{figure}

As mentioned before, PACS can be integrated into Merge search in a straightforward manner. This integration, which was explained before, is graphically shown in the flow chart of  Figure~\ref{fig:integration}.

This flowchart shows one run of Algorithm~\ref{alg:ms} with PACS. The first step is to generate the $n_s$ solutions, which is done in the {\it PACS} step. In particular, if we have five colonies, we generate five different solutions. This solution pool (in addition to the best-so-far solution) is used to generate the partitions or sets of aggregated variables ({\it Generate Partitions}). These partitions are further split using random splitting in step {\it Random Split}. All the prior steps lead to defining the restricted MIP ({\it Generate R-MIP}), which is then solved in step {\it Solve R-MIP}. The above steps constitute the four steps of Merge search, which are then repeated until the time limit is expired.

\section{Experimental Setting}\label{sec:expts}

We implemented Merge search in C++ compiled in GCC-5.4.0. The MIPs were solved with Gurobi 8.1.0\footnote{\url{http://www.gurobi.com/}}. For executing multiple colonies in parallel OpenMP \citep{Dagum98} was used. All experiments were carried out on Monash University's Campus Cluster, MonARCH,\footnote{\url{https://confluence.apps.monash.edu/display/monarch/MonARCH+Home}}. The machines on MonARCH consist of 24 cores and 256 GB RAM. Each physical core consists of two hyper-threaded cores with Intel Xeon E5-2680 v3 2.5GHz, 30M Cache, 9.60GT/s QPI, Turbo, HT, 12C/24T (120W).

\subsection{Problem Instances}

For the experimental evaluation we make use of the benchmark set that was already used in the original study by \citet{kimms00}. These instances can be found in PSPLIB \citep{KOLISCH1997205}.\footnote{\url{https://www.sciencedirect.com/science/article/abs/pii/S0377221796001701}} 

The instances from PSPLIB cover a range of characteristics, which are designed to examine different aspects of the RCPS problem. The different characteristics include the number of tasks $\{30, 60, 90, 120\}$, resource factors $\{$0.25, 0.5, 0.75, 1.0$\}$, network complexities $\{$1.5, 1.8, 2.1$\}$, and resource strengths. Problem instances consisting of 30, 60 and 90 tasks are generated with resource strengths from $\{$0.2, 0.5, 0.7, 1.0$\}$, whereas the problem instances with 120 tasks have resource strengths from $\{$0.1, 0.2, 0.3, 0.4, 0.5$\}$.  The resource strength indicates the scarceness of resources, where low values imply very tight resource constraints. The resource factor specifies the total amount of each resource used by a task, as a proportion of the total amount available of the resource. The network complexity specifies the amount of precedence constraints in a problem instance, with a large value implying more precedence constraints. The datset consist of a large number of problem instances, each one obtained by a combination of the four characteristics discussed above. In particular, there are 10 instances for each combination, leading to 480 problem instances for 30, 60, and 90 tasks, and 600 problem instances for 120 tasks. Hence, there are a  total of 2040 problem instances in the dataset.

The cash flows for all tasks in a dataset are obtained by using the study by~\citet{kimms00} as a guide. Each task is considered independently. The latest start time ($l_i$) for this task  $l_j = \sum_i d_i$ $\forall i \rightarrow j$ (i.e. the sum of the durations of the predecessors of task $j$), and then $\delta := 3.5 \times \max_j l_j$. A task's cash flow $c_i$ is generated uniformly at random from $\left[-500, 1000 \right]$. The discount rate is determined as $\alpha := \sqrt[52]{1 + 0.05} - 1$, the motivation for which is provided in \citep{kimms00}. 

Preliminary experiments were conducted to tune Merge search. This analysis is presented in Appendix~A. Merge search, along with know state-of-the-art methods, are analysed in the following section.

\section{Results} \label{sec:results}
We investigate how MS-PACS performs for the RCPS-NPV problem. We compare MS-PACS to Construct, Solve, Merge, Adapt (CMSA) and Lagrangian relaxation with ant colony optimisation (LRACO) (the two current state-of-the-art methods) and to the standalone version of PACS. In order to make fair comparisons, we run all algorithms on the same resource---MonARCH---as discussed earlier. 

Unless otherwise specified, we allow 5 cores per run. This means that, for MS-PACS, the MIP solver (Gurobi) can use up to 5 cores and there are up to 5 colonies used in PACS.\footnote{Each individual colony uses the same parameter settings as obtained from \citep{Thiruvady2014ps}.} Additionally, to make fair comparisons, the CMSA, LRACO and PACS implementations are also given 5 cores. Each run was given 900 seconds of wall clock time.

Due to the complexity of MS-PACS and its associated components (especially PACS), we also analyse the impact of different components of the algorithm. Specifically, we investigate (a) performance characteristics with varying the number of cores and (b) varying random splitting. 

\subsection{Comparing MS-PACS, CMSA, LRACO and PACS}

\begin{table}[t]
\centering
\caption{Average gaps of MS-PACS, CMSA, LRACO and PACS to the upper bound obtained by LRACO computed as $\left(\frac{UB-LB}{UB}\right)\times 100$. Statistically significant results at $\alpha = 0.05$ are highlighted in boldface and italics. The number of instances for which an algorithm finds the best solution is shown in column {\it \# best}; MS-PACS = Merge Search with parallel ACS, CMSA: Construct, Solve, Merge and Adapt with parallel ACS, LRACO: Lagrangian relaxation and ant colony optimisation, PACS: Parallel ACS.}
\label{tab:comp}
\scalebox{0.8}{
\begin{tabular}{|lc|ccc|ccc|ccc|ccc|}
\hline										
Tasks	&	&	\multicolumn{3}{c|}{MS-PACS}	&	\multicolumn{3}{c|}{CMSA}	& \multicolumn{3}{c|}{LRACO}	& \multicolumn{3}{c|}{PACS}	\\
&&	Mean	&	SD	&	\# best	&	Mean	&	SD	&	\# best	&	Mean	&	SD	&	\# best & Mean	&	SD	&	\# best \\ 
\hline
30	&&		1.567		&	1.985	&	{\bf	396}	&	1.567	&	1.984	&	{\bf	396}	&	1.634	&	2.023	&	225	&	1.616	&	2.023	&	240	\\
60	&&	{\bf \it	1.329}	&	1.177	&	{\bf	355}	&	1.334	&	1.179	&		269	&	1.483	&	1.251	&	110	&	1.469	&	1.263	&	114	\\
90	&&	{\bf \it	1.339}	&	1.020	&	{\bf	348}	&	1.362	&	1.033	&		209	&	1.472	&	1.049	&	114	&	1.552	&	1.116	&	0	\\
120	&&	{\bf \it	1.634}	&	1.059	&	{\bf	213}	&	1.653	&	1.054	&		185	&	1.763	&	0.987	&	84	&	2.141	&	1.217	&	0	\\

\hline
\end{tabular}
}
\end{table}

The first comparison of all three methods is on the problem instances from PSPLIB. Table~\ref{tab:comp} shows the results of all three algorithms separated by problem size, i.e.~the number of tasks. The table reports average gaps, which are obtained as the gap for each algorithm (lower bound or LB) to the upper bound (UB) of LRACO: $\frac{UB-LB}{UB}$.\footnote{We use the LRACO upper bounds for this purpose, because previous studies have shown that these upper bounds are relatively tight.} We see that, with respect to NPV, MS-PACS is the best performing method on average. For the smallest problem instances (30 tasks), MS-PACS and CMSA are relatively close and while MS-PACS is slightly more effective, this is not a statistically signifcant results. However, for 60, 90 and 120 tasks MS is more effective statistically speaking and the differences between the algorithms increase with problem size. Additionally, for the problem instances with 60, 90 and 120 tasks, LRACO and PACS are consistently worse than both MS-PACS and CMSA. 

Regarding the number of times the best solution was found for each problem size (column with heading '$\#$ best'), we see that for 30 tasks MS-PACS and CMSA obtain the same results. However, with an increasing number of tasks, MS-PACS is clearly the best performing method. LRACO and PACS are effective for small problems, however, PACS in particular does not scale well. In fact, for the instances with 90 and with 120 tasks, PACS never finds the best solution for any instance. \\

\begin{table}[ht]
\centering
\caption{Breakdown of the results for MS-PACS and CMSA by resource factor, resource strength and network complexity. Standard deviations are shown in parentheses.} 
\label{tab:breakdown}
\scalebox{0.8}{
\begin{tabular}{|l|c|cccc|}
\hline
\hline
\multicolumn{6}{|c|}{MS-PACS}																					\\
\hline
\hline
	&		&	30				&	60				&	90				&	120				\\
	&	1.50	&	{\bf	1.662}	(1.92)	&	{\bf	1.362}	(1.28)	&	{\bf	1.460}	(1.25)	&	{\bf	1.714}	(1.19)	\\
NC	&	1.80	&	{\bf	1.399}	(1.73)	&	{\bf	1.412}	(1.20)	&	{\bf	1.254}	(0.88)	&	{\bf	1.560}	(0.95)	\\
	&	2.10	&		1.640		(2.27)	&	{\bf	1.213}	(1.03)	&	{\bf	1.302}	(0.88)	&	{\bf	1.629}	(1.03)	\\
\hline																															
	&	0.25	&	{\bf	1.872}	(2.62)	&	{\bf	1.295}	(0.96)	&	{\bf	1.101}	(0.88)	&	{\bf	1.176}	(0.69)	\\
RF	&	0.50	&		1.435		(1.68)	&	{\bf	1.377}	(1.46)	&	{\bf	1.418}	(0.89)	&	{\bf	1.765}	(1.17)	\\
	&	0.75	&		1.339		(1.34)	&	{\bf	1.313}	(1.14)	&	{\bf	1.395}	(1.16)	&	{\bf	1.957}	(1.16)	\\
	&	1.00	&	{\bf	1.621}	(2.06)	&	{\bf	1.331}	(1.10)	&	{\bf	1.441}	(1.09)	&	{\bf	1.640}	(0.99)	\\
\hline																															
	&	0.10	&							&							&							&		2.344		(1.12)	\\
	&	0.20	&		1.395		(0.87)	&	{\bf	1.445}	(0.89)	&	{\bf	1.633}	(0.62)	&	{\bf	1.722}	(1.04)	\\
	&	0.30	&							&							&							&	{\bf	1.613}	(0.97)	\\
RS	&	0.40	&							&							&							&	{\bf	1.317}	(0.64)	\\
	&	0.50	&		1.366		(2.17)	&	{\bf	1.192}	(1.06)	&	{\bf	1.205}	(0.86)	&	{\bf	1.176}	(1.07)	\\
	&	0.70	&	{\bf	1.489}	(2.18)	&	{\bf	1.207}	(1.19)	&	{\bf	1.018}	(0.78)	&							\\
	&	1.00	&		2.021		(2.33)	&		1.469		(1.47)	&		1.494		(1.49)	&							\\
\hline
\hline
\multicolumn{6}{|c|}{CMSA}	\\
\hline
\hline
	&		&	30				&	60				&	90				&	120				\\
	&	1.50	&		1.663		(1.92)	&		1.369		(1.28)	&		1.483		(1.25)	&		1.722		(1.17)	\\
NC	&	1.80	&		1.401		(1.73)	&		1.415		(1.20)	&		1.286		(0.91)	&		1.577		(0.93)	\\
	&	2.10	&	{\bf	1.636}	(2.27)	&		1.220		(1.04)	&		1.316		(0.89)	&		1.659		(1.05)	\\
\hline																															
	&	0.25	&		1.873		(2.62)	&		1.297		(0.97)	&		1.106		(0.89)	&		1.183		(0.69)	\\
RF	&	0.50	&	{\bf	1.434}	(1.68)	&		1.383		(1.46)	&		1.430		(0.90)	&		1.769		(1.17)	\\
	&	0.75	&		1.339		(1.34)	&		1.315		(1.14)	&		1.427		(1.18)	&		1.983		(1.15)	\\
	&	1.00	&		1.622		(2.06)	&		1.344		(1.11)	&		1.482		(1.10)	&		1.676		(0.98)	\\
\hline																															
	&	0.10	&							&							&							&	{\bf	2.339}	(1.07)	\\
	&	0.20	&	{\bf	1.393}	(0.86)	&		1.461		(0.89)	&		1.702		(0.67)	&		1.767		(1.06)	\\
	&	0.30	&							&							&							&		1.641		(0.98)	\\
RS	&	0.40	&							&							&							&		1.329		(0.65)	\\
	&	0.50	&	{\bf	1.367}	(2.17)	&		1.196		(1.07)	&		1.219		(0.87)	&		1.187		(1.07)	\\
	&	0.70	&		1.490		(2.18)	&		1.209		(1.20)	&		1.025		(0.78)	&							\\
	&	1.00	&		2.021		(2.33)	&		1.469		(1.47)	&		1.494		(1.49)	&							\\
\hline
\end{tabular}
}
\end{table}

We now break the results down to the different factors associated with the problem instances. Table~\ref{tab:breakdown} shows the results of MS-PACS and CMSA on all problem instances summarised by resource factor, resource strength and network complexity. We see that, consistently, MS-PACS outperforms CMSA except when the resource strength is 1.0. Here, both methods generate exactly the same results. The corresponding results for LRACO are presented in Appendix~\ref{app:lraco}. They are not shown here since MS-PACS and CMSA consistently outperform LRACO.

\subsection{Investigating Parallelisation}
\label{sec:parallelism}

A key aspect of the hybrids of Merge search and CMSA with PACS is their ability to efficiently utilise whatever resource is available to them. Throughout the execution of both algorithms we can expect to see a core utilisation of nearly 100$\%$, using any number of cores. This is because in PACS any number of colonies can be executed in parallel (using as many colonies as cores) and the MIP solvers also efficiently use any number of cores provided (a feature of modern commercial integer programming solvers). Nevertheless, we aim to determine how the number of cores effects the performance, and in particular, the convergence properties of the algorithms.
 
\begin{figure*}[t!]
  \includegraphics[width=16cm]{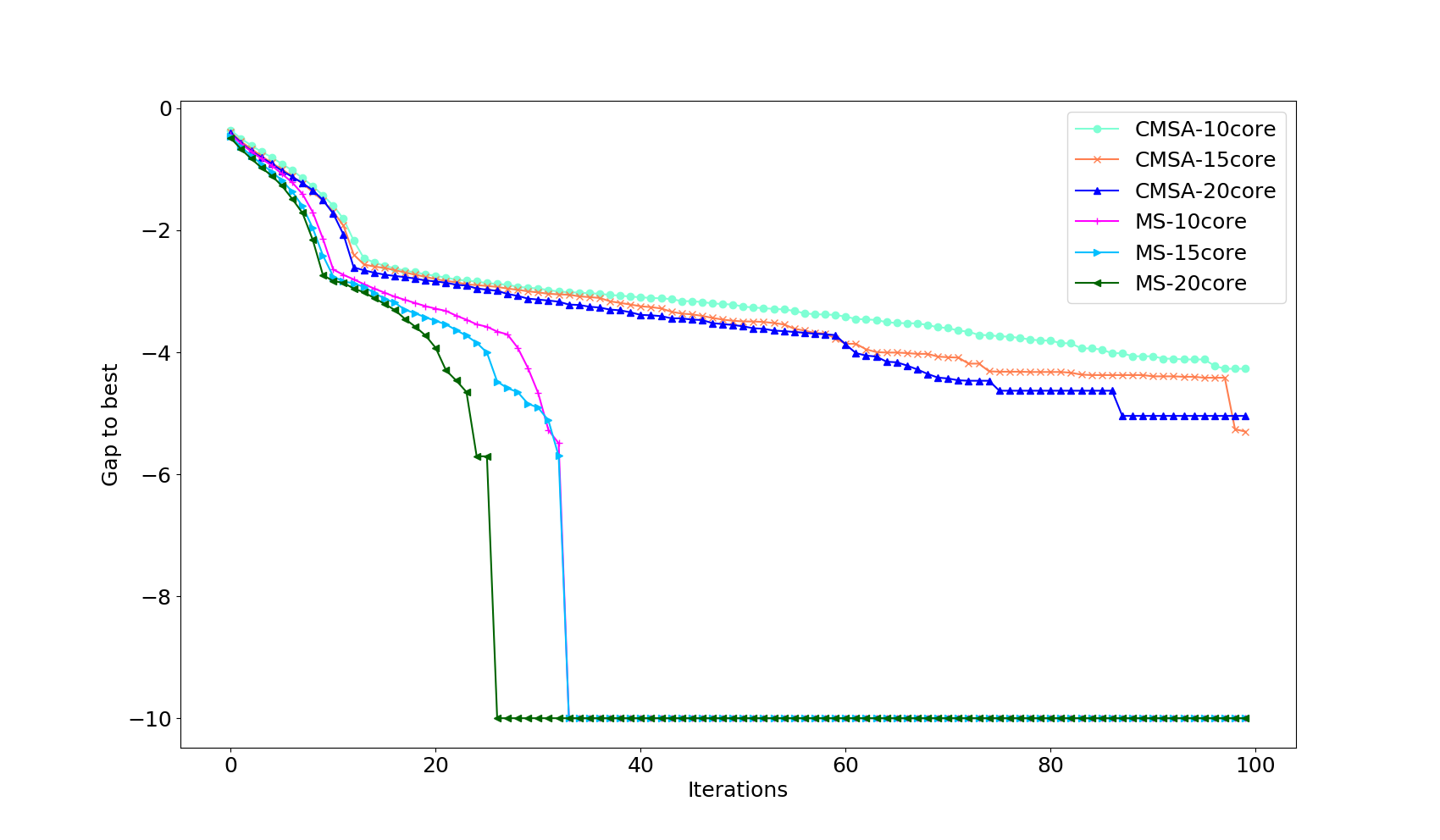}
  \centering
  \caption{The performance of MS-PACS and CMSA with a varying number of cores. The results are obtained by considering all instances with 120 tasks and averaging over the percentage difference to the best solution ($\frac{(UB-Best)\times100}{Best}$). The y-axis is log-scaled.}
  \label{fig:convergence}
\end{figure*}

Figure~\ref{fig:convergence} shows how MS-ACS and CMSA progress in their respective searches. For this experiment, we consider all problem instances with 120 tasks and execute both MS-PACS and CMSA with 10, 15 and 20 cores. For the PACS component, we allowed the same number of colonies as the number of available cores (only one of which is seeded with the best-so-far solution). For each problem instance, we consider the best solution found by any algorithm and compute the percentage difference from the current incumbent to this best solution ($\frac{\left(UB-Best\right)\times100}{Best}$). The percentage differences are then averaged for each iteration and for each problem instance, and we plot the progress for 100 iterations in the graphic. 

There are two main results to observe in Figure~\ref{fig:convergence}. Firstly, and mainly, the MS-PACS versions (characterised by the number of utilised cores) exhibits a much faster convergence than the corresponding CMSA version to the best solutions. In fact, with an increasing number of iterations, the algorithms diverge in their performance with MS-PACS achieving substantial gains. 

Secondly, considering only MS-PACS, we see that with an increasing number of cores, the time required to reach the best solution generally decreases. In particular, using 20 cores is faster than using 15 cores, which---in turn---is faster than using 10 cores. This is also the case for CMSA. While this is not entirely surprising, this demonstrates that both MS-PACS and CMSA scale well when provided with a large number of resources. Arguably, this effect is more pronounced in the case of MS-PACS.
  
\section{Conclusion}\label{sec:conclusion}

This study has introduced the application of a recent matheuristic technique, Merge search, to the resource constrained project scheduling problem with the aim of maximising the net present value, the RCPS-NPV problem. Moreover, we compared our approach to another matheuristic technique, CMSA, recently presented for the same problem. Using well-known benchmark problem instances for the RCPS-NPV problem, we find that MS-PACS is clearly the more effective approach. In particular, with increasing problems sizes (up to 120 tasks), Merge search shows increasing gains compared to CMSA, which in turn, both outperform PACS on its own. An investigation of the convergence properties of Merge search and CMSA shows that Merge search is indeed able to converge to higher quality solutions a lot more quickly than CMSA. 

A key aspect of future work will be to consider how Merge search and CMSA scale in solving large-scale problems. In the current study, we investigate benchmark instances of up to 120 tasks. However, the scaling of the method to consider realistic problem sizes of up to 1000 tasks could prove very interesting. 

We would like to emphasize that Merge search allows the easy integration of a MIP solver with (parallel) ACS. If a problem can be efficiently modeled as a MIP or solved by ACS, the proposed approach provides a straightforward enhancement, which can more effectively solve the original problem. Moreover, multi-core shared memory architectures are very common nowadays and we see that the current proposed architecture is able to efficiently utilise these resources through an entire run. The parallel framework allows maintaining a good trade-off between intensification and diversification, typically required by metaheuristic approaches~\citep{blumroli03}. 

Another direction concerning parallel implementations could be to investigate a distributed framework with Merge search and PACS. We are currently exploring the use of a highly-parallel framework across multiple nodes, and the results show that for very large open pit mining problems, very good solutions can be found via message passing interface (MPI) based parallelisations. In a similar fashion, it will be interesting to see how a highly-parallel MS-ACS will perform on the project scheduling problem instances studied here and also how the method scales with larger project scheduling problem instances.

This study has demonstrated that Merge search is effective when using a time-based MIP formulation. However, alternative formulations might be even more suitable. Possible alternatives include sequence-based formulations where the decision variables are associated with sequences of tasks. Devising an effective merge step will be key to obtaining an efficient approach in these cases.

\section*{Acknowledgements}
 This research was supported in part by the Monash eResearch Centre and eSolutions-Research Support Services through the use of the MonARCH HPCC luster.

\bibliographystyle{elsarticle-harv}
\bibliography{references}

\appendix
\section{Merge search Parameter Selection} \label{app:cmsa_params}

Similar to the study by \citet{Thiruvady2019}, we consider a subset of the data available from PSPLIB for parameter tuning experiments. The problem instances are chosen considering the number of tasks (60 and 120), resource factors (0.5 and 1.0) and strengths (0.5 and 1.0). An individual ACS colony's settings was the same as that of \citep{Thiruvady2014ps}. For each run, five cores were provided, allowing five colonies to to be run in parallel.

For MS-PACS, the parameters of interest are the MIP time limit ($t_{iter}$) and the number of split sets ($K$), whereas for ACS we only consider the number of iterations allowed for each colony. The MIP time limit is set to either 60 or 120 seconds, ensuring that there is sufficient time for the MIP solver to find good solutions. At the same time these time limits do not use up too much of the total running time. The number of split sets ($K$) is of particular interest. To test a wide range of values, we chose 10, 50, 100, 200 and 500. In principle, we could test values larger than 500. However, in preliminary experimentation we found that most of the generated MIPs cannot be solved to optimality in this case. The total number of ACS iterations are limited to 500, 1000 and 2000.\footnote{Note that the number of colonies---respectively, the number of cores---is also a parameter of interest, but we analyse this aspect of the algorithm in Section~\ref{sec:parallelism}.}

\begin{table}[h!]
\centering
\caption{By varying parameter values of MS-PACS, the following number of best solutions were found for each parameter respectively; $t_{iter}$ - MIP time limit; ACS iter. - ACS iterations; $K$ - split sets.}
\label{tab:cmsa_params_1}
\scalebox{0.8}{
\begin{tabular}{|l|c|cc|}
\hline
Param.	&	Value	&	60 tasks	&	120 tasks	\\
\hline
$t_{iter}$	&	60	&	{\bf	10}	&	{\bf	8}	\\
	&	120	&		0		&		4		\\
\hline
	&	500	&	6	&		4		\\
ACS Iter.	&	1000	&		9		&		2		\\
	&	2000	&		{\bf	9}		&	{\bf	4}		\\
\hline
	&	10	&	6	&		0		\\
	&	50	&	8	&	    0	\\
$K$	&	100	&	7	&		1		\\
    &	200	&	8	&		2		\\
    &	500	&	{\bf	8}	&	{\bf	7}		\\
\hline
\end{tabular}
}
\end{table}

For each of the settings, Table~\ref{tab:cmsa_params_1} shows the number of runs in which the best solution was found (note, that if two settings achieve the same NPV value, both solutions are included here). We proceed by considering one parameter, fixing it while varying the second parameter and then fix the first and second parameters while varying the third. For the ordering, we choose the MIP time limit followed by the number of ACS iterations and finally, the number of split sets. As we see in these results, a MIP time limit of 60 seconds is clearly the best option. Concerning the number of ACS iterations, 2000 iterations is the best option. Finally, the best number of split sets is 500.

\section{Performance of LRACO on the PSPLIB Problem Instances}
\label{app:lraco}

\begin{table}[ht]
\centering
\caption{Breakdown of the results by resource factor, resource strength and network complexity for LRACO. Standard deviations are shown in parentheses.}
\label{tab:breakdownLRACO}
\scalebox{0.8}{
\begin{tabular}{|l|c|c|c|c|c|}
\hline
	&	Tasks	&	30				&	60				&	90				&	120				\\
\hline
	&	1.50	&	1.738	(1.93)	&	1.529	(1.32)	&	1.600	(1.27)	&	1.836	(1.14)	\\
NC	&	1.80	&	1.463	(1.74)	&	1.571	(1.32)	&	1.394	(0.92)	&	1.678	(0.81)	\\
	&	2.10	&	1.704	(2.35)	&	1.348	(1.11)	&	1.423	(0.92)	&	1.774	(0.98)	\\
\hline												
	&	0.25	&	1.878	(2.62)	&	1.361	(1.03)	&	1.212	(0.92)	&	1.475	(0.78)	\\
RF	&	0.50	&	1.482	(1.69)	&	1.558	(1.56)	&	1.612	(0.97)	&	1.926	(1.19)	\\
	&	0.75	&	1.431	(1.38)	&	1.494	(1.22)	&	1.533	(1.18)	&	2.005	(1.05)	\\
	&	1.00	&	1.748	(2.18)	&	1.518	(1.15)	&	1.534	(1.08)	&	1.646	(0.79)	\\
\hline							
	&	0.10	&					&					&					&	2.404	(0.99)	\\
	&	0.20	&	1.551	(0.97)	&	1.764	(1.08)	&	1.832	(0.69)	&	1.840	(0.93)	\\
	&	0.30	&					&					&					&	1.721	(0.83)	\\
RS	&	0.40	&					&					&					&	1.454	(0.58)	\\
	&	0.50	&	1.448	(2.27)	&	1.401	(1.15)	&	1.423	(0.92)	&	1.397	(1.17)	\\
	&	0.70	&	1.520	(2.19)	&	1.291	(1.23)	&	1.134	(0.80)	&					\\
	&	1.00	&	2.021	(2.33)	&	1.469	(1.47)	&	1.494	(1.49)	&					\\
\hline

\end{tabular}
}
\end{table}
\end{document}